\documentclass[letterpaper]{article} 
\usepackage[preprint]{aaai2027}
\usepackage[hyphens]{url}  
\usepackage{graphicx} 
\usepackage{natbib}  
\usepackage{caption} 
\usepackage{algorithm}
\usepackage{algorithmic}

\usepackage{caption} 
\usepackage[table]{xcolor}

\usepackage[most]{tcolorbox}
\usepackage{tabularx}
\usepackage{enumitem}
\usepackage{booktabs}
\usepackage{amsmath}
\usepackage{amssymb}
\usepackage{enumitem}
\usepackage{subfigure}		
\usepackage{multirow}
\usepackage{booktabs}
\usepackage{float}    
\usepackage{newfloat}
\usepackage{listings}
\DeclareCaptionStyle{ruled}{labelfont=normalfont,labelsep=colon,strut=off} 
\floatstyle{ruled}
\newfloat{listing}{tb}{lst}{}
\floatname{listing}{Listing}

\usepackage{booktabs}

\title{ReflectFact: Self-Reflective Agents for Improving Comprehension and Reasoning in Multi-Hop Fact Verification}
\author{
    Runze Zhao\textsuperscript{\rm 2}, Zixin Tang\textsuperscript{\rm 1}\corresponding, Xiaoshuai Hao\textsuperscript{\rm 3}, Leyuan Chang\textsuperscript{\rm 1},
    Xiaopeng Fu\textsuperscript{\rm 1}, Boyu Qiao\textsuperscript{\rm 2},
    Dongyang Zhang\textsuperscript{\rm 1}
}

\affiliations{
    \textsuperscript{\rm 1}Zhongguancun Laboratory \quad
    \textsuperscript{\rm 2}Institute of Information Engineering, Chinese Academy of Sciences \quad
    \textsuperscript{\rm 3}Xiaomi EV
}

\begin{document}

\maketitle

\begin{abstract}
Multi-hop fact verification, which verifies claims by reasoning over multiple pieces of evidence, is critical for combating misinformation on social media yet remains highly challenging. Recent methods primarily rely on multi-agent collaboration to decompose fact verification into specialized subtasks. 
However, these methods face two critical limitations: (1) agents may perform individual subtasks without sufficient awareness of the global verification objective, causing their reasoning to deviate from the intended direction; and (2) conflicts between parametric knowledge and the provided evidence may undermine evidence-grounded reasoning and lead to incorrect verdicts. To address these challenges, we propose \textbf{\textit{ReflectFact}}, a novel self-reflective agent framework for multi-hop fact verification. ReflectFact introduces three key tasks. \textbf{\textit{Explicit Reasoning Path Planning}} builds an evidence-grounded reasoning path by resolving implicit entities, decomposing the claim into sub-questions, and integrating the verified facts into a verdict. \textbf{\textit{Evidence-Drift Verification}} makes the agent re-answer by quoting the supporting evidence when a grounded answer merely echoes its parametric prior, thereby calibrating evidence deviation to ensure grounded comprehension. \textbf{\textit{Reasoning Reflection Verification}} re-examines each reasoning step and regenerates it once an inconsistency is detected, correcting reasoning flaws such as location bias and replacement bias through a global task perspective. Subsequently, the agent aggregates validated reasoning chains to yield reliable verdicts. Extensive experiments on HOVER and EX-FEVER demonstrate that ReflectFact effectively remedies the comprehension and reasoning defects of existing methods, achieving state-of-the-art performance and respectively outperforming the strongest baseline by 3.32\% and 2.78\% on the two datasets.
\end{abstract}

\section{Introduction}
The pervasive dissemination of misinformation on social media has driven an urgent need for automated fact verification \cite{park2021faviq,si2023exploring,liu2023keshem,wang-etal-2024-factcheck}. Multi-hop fact verification, which assesses claim veracity through multi-step reasoning over multiple pieces of evidence \cite{ma2023ex,zhu2023explain,pan2023fact,10.1609/aaai.v39i1.32034}, more closely mirrors real-world misinformation scenarios and has thus attracted substantial research attention. As a result, developing robust multi-hop verification systems significantly impacts the reliability and trustworthiness of information ecosystems in practical applications.

\begin{figure}[!t]
  \centering
  \subfigure[Objective conflicts.]{\includegraphics[width=0.47\linewidth]{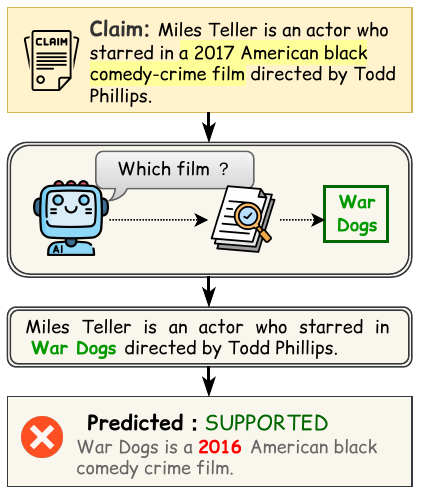}}
  \hfill
  \subfigure[Knowledge conflicts.]{\includegraphics[width=0.47\linewidth]{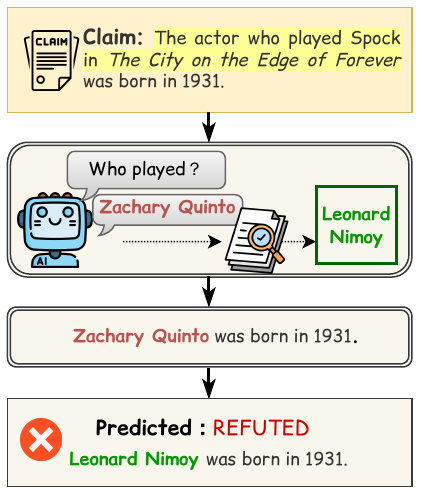}}
  \caption{Examples illustrating objective conflicts and knowledge conflicts in existing fact verification methods, highlighting their limitations in multi-hop fact verification.}
  \label{fig_1}
  \vspace{-0.5em}
\end{figure}

Recently, the rapid development of large language models (LLMs) has significantly advanced fact verification capabilities~\cite{pan2023fact, zhang2023towards, yue2023metaadapt, si-etal-2024-checkwhy}, enabling more sophisticated multi-hop reasoning approaches. Researchers have explored three main categories of solutions: knowledge graphs, fine-tuned natural language inference (NLI) models, and agent-based approaches. Early methodologies employ knowledge graphs or fine-tuned NLI models to verify multi-hop facts~\cite{ren2020beta, hedebertav3, lei2025factcg}, effectively capturing logical dependencies through structured representations. Recent approaches leverage intelligent agents that mimic human fact-verification workflows~\cite{yang2026verdict, zhang2023towards, xu2025multimodal, wang-etal-2024-factcheck}, decomposing verification into sequential subtasks with progressive execution. Despite their promise, agent-based methods still face two critical limitations, as illustrated in Figure~\ref{fig_1}. \textit{(1)Objective conflicts.} Agents performing individual subtasks may lack sufficient awareness of the global verification objective, causing their reasoning to deviate from the intended direction. As shown in Figure~\ref{fig_1}(a), the entity resolution agent directly replaces the descriptive mention ``a 2017 American black comedy crime film'' with the actual film title ``War Dogs''. Although this substitution satisfies the local entity resolution objective, it overlooks the inconsistency between the stated release year and the film’s actual release year, thereby obscuring critical evidence for fact verification. \textit{(2) Knowledge conflicts.} Conflicts may arise when the parametric knowledge in an agent is inconsistent with evidential knowledge and is referred to as evidence drift. As illustrated in Figure~\ref{fig_1}(b), it can lead to unsupported modifications to an otherwise valid claim. Therefore, we propose ReflectFact, a self-reflective agent framework with post-verification that align local reasoning with the global objective and ensure evidence-grounded verification.

In this paper, we propose \textbf{\textit{ReflectFact}}, a novel self-reflective agent framework to address limitations of existing agent-based methods in multi-hop fact verification through self-reflective post-hoc verification. 
ReflectFact introduces three tasks coordinated across verification stages. \textbf{\textit{Explicit Reasoning Path Planning}} builds an evidence-grounded reasoning path through three automated pipelines: \textit{Implicit Entity Resolution} identifies implicitly referenced entities, \textit{Semantic Decomposition} breaks claims into atomic sub-claims, and \textit{Integrative Logical Reasoning} constructs coherent logical chains for verdicts. 
Unlike standard workflows that execute subtasks sequentially without verification, ReflectFact ensures comprehensive verification coverage. \textbf{\textit{Evidence-Drift Verification}} ensures agents prioritize grounded evidence over parametric knowledge, making the agent re-answer by quoting the supporting evidence to prevent knowledge conflicts. \textbf{\textit{Reasoning Reflection Verification}} provides agents with global information to re-examine after each step. 
Leveraging the observation that LLMs exhibit stronger verification than generation capabilities, it facilitates self-evaluation of reasoning coherence with the overarching objective, thereby mitigating biases arising from sub-task optimization. The synergy between structured reasoning-path construction, evidence comprehension, and reflective verification enables robust adaptation to diverse multi-hop scenarios where error accumulation and knowledge bias are prevalent. 
Extensive experiments on HOVER~\cite{jiang2020hover} and EX-FEVER~\cite{ma2023ex} demonstrate that ReflectFact consistently outperforms the strongest of ten competitive baselines by 3.32\% and 2.78\% in overall Macro-F1, achieving SOTA with robust scalability and interpretability.

Our main contributions are summarized as follows:
\begin{itemize}

    \item We identify two key limitations faced by agent-based methods—namely, a lack of a global reasoning perspective and an over-reliance on parametric knowledge—which render them prone to both objective conflicts and knowledge conflicts.

    \item We propose \textbf{ReflectFact}, a novel self-reflective agent framework for multi-hop fact verification, introducing self-reflective post-hoc verification at each reasoning step to address the existing limitations in subtask-level reasoning quality, ensuring accurate verification in complex multi-hop scenarios.  
  
    \item \textbf{ReflectFact} achieves significant performance gains on HOVER and EX-FEVER, outperforming the strongest of ten competitive baselines and demonstrating superior adaptability in handling complex fact verification.
\end{itemize}


\section{Related Work}
\textbf{Fact Verification.}
Fact verification predicts claim veracity given retrieved evidence~\cite{park2021faviq,botnevik2020brenda,si2023exploring,liu2023keshem,cekinel-etal-2025-multimodal}. Early approaches encode claims with textual evidence for classification, such as entity linking with enhanced language models~\cite{hanselowski-etal-2018-ukp} and BERT-based explainable verification~\cite{kotonya2020explainable}, or reason over graphs built from claims and evidence, such as semantic-level graph reasoning~\cite{zhong-etal-2020-reasoning} and kernel graph attention networks~\cite{liu-etal-2020-fine}. However, these single-hop methods struggle when verdicts depend on jointly combining facts from multiple sources. To address multi-hop scenarios, recent methods improve reasoning through counterfactual data augmentation~\cite{zhu2023explain} and LLM-extracted structured knowledge injection~\cite{10.1609/aaai.v39i22.34520}. However, these approaches primarily focus on evidence retrieval and knowledge augmentation while overlooking reasoning quality at each intermediate step, leading to error accumulation during multi-hop inference.
To improve multi-hop reasoning, Zhu et al.~\cite{zhu2023explain} generate additional data through a counterfactual procedure to enhance generalization, and \cite{10.1609/aaai.v39i22.34520} inject LLM-extracted structured knowledge to strengthen multi-hop verification. 
However, these methods mainly rely on predefined representations or external knowledge structures, which limits their ability to flexibly reason over complex claims involving implicit relations and multiple evidence sources.
These methods mainly improve evidence representation, training data, or relation modeling, rather than explicitly verifying the intermediate outputs produced during multi-hop fact verification.

\textbf{LLMs for Fact Verification.}
The strong understanding and reasoning abilities of LLMs have opened new opportunities for fact verification \cite{zhang2023towards,pan2023fact,si-etal-2024-checkwhy}. One line of work elicits step-by-step reasoning through prompting, such as Chain-of-Thought \cite{wei2022chain} and Tree-of-Thought \cite{yao2024tree}. A second line decomposes a claim into sub-problems: HiSS \cite{zhang2023towards} partitions a claim and verifies each part, ProgramFC \cite{pan2023fact} prompts LLMs to generate reasoning programs from a shared function library, and Factcheck-GPT \cite{wang-etal-2024-factcheck} builds a fine-grained checking pipeline. A third line augments LLMs with external structure or collaboration, \textit{e.g.}, multiple role-playing agents \cite{10.1609/aaai.v39i1.32034} and knowledge graphs extracted from the input \cite{chen-etal-2025-graphcheck}.
However, these approaches largely rely on the correctness of generated reasoning trajectories and lack mechanisms to verify whether intermediate reasoning steps remain consistent with the evidence and the overall verification objective.
\begin{figure*}[t!]
    \centering
    \includegraphics[width=\textwidth]{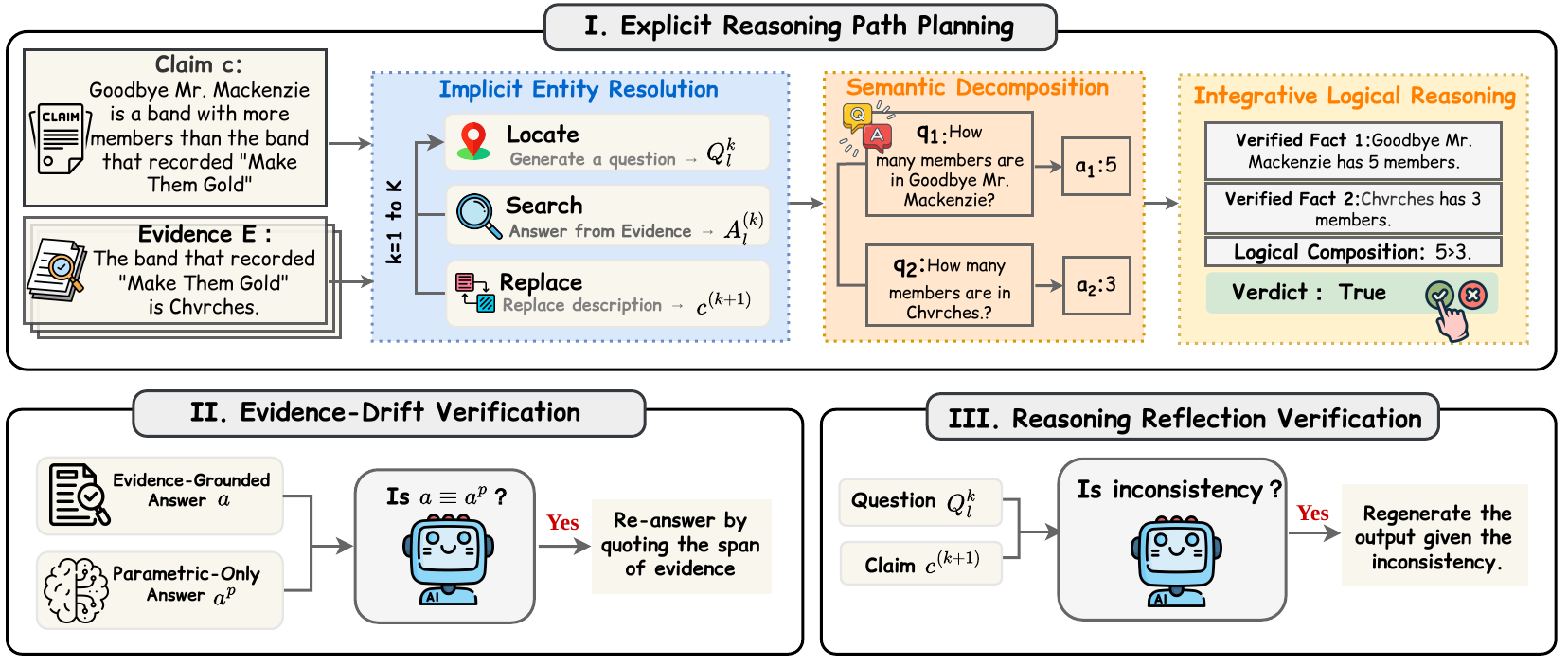}
    \caption{\textbf{Overview of ReflectFact.} ReflectFact is divided into three tasks. \textbf{Explicit Reasoning Path Planning} builds an evidence-grounded reasoning path by resolving implicit entities, decomposing the claim into sub-questions, and integrating the verified facts into a verdict. \textbf{Evidence-Drift Verification} makes the agent re-answer by quoting the supporting evidence when a grounded answer merely echoes its parametric prior. \textbf{Reasoning Reflection Verification} re-examines each reasoning step and regenerates it once an inconsistency is detected.}
    \label{method}
\end{figure*}

\textbf{Agent-based Fact Verification.}
Agent-based systems further organize verification as coordinated workflows.
BiDeV integrates multiple role-played LLMs and performs bilateral defusing to resolve claim vagueness and filter redundant evidence~\cite{10.1609/aaai.v39i1.32034}.
AgentFact coordinates specialized agents for strategy planning, evidence retrieval, visual analysis, reasoning, and explanation generation in multimodal fact checking~\cite{xu2025multimodal}.These methods decompose verification into subtasks and execute them through agent workflows. These methods decompose fact verification into subtasks and execute them through agent workflows. However, existing agent-based methods primarily focus on task decomposition and workflow orchestration, while paying limited attention to the reliability of reasoning within individual subtasks, leaving the overall process vulnerable to objective and knowledge conflicts. In contrast, ReflectFact applies post-verification after each subtask, using reflective reasoning to evaluate intermediate outputs against the global fact-verification objective and the available evidence.


\section{Methodology}
\subsection{Task Definition}

Multi-hop fact verification can be formulated as follows. Given a claim $c$ and a set of relevant evidence $E={e_1,e_2,\ldots,e_n}$ drawn from a large textual corpus such as Wikipedia, the goal is to determine whether the claim is supported or refuted through multi-step reasoning over the evidence. We denote the final verification label by $y \in \{True, False\}$. Since our task focuses on evidence comprehension and reasoning rather than evidence retrieval, we use dataset-provided gold evidence, where each $e_i\in E$ contributes to supporting or refuting the claim.

\subsection{Explicit Reasoning Path Planning}

We design the execution pipeline of the agent framework to construct an explicit, evidence-grounded reasoning path through step-by-step multi-hop claim verification. Furthermore, we introduce two self-reflective verification tasks, \textit{Evidence-Drift Verification} and \textit{Reasoning Reflection Verification}, to suppress errors that arise during the execution of sub-tasks.
We first describe how \textit{Explicit Reasoning Path Planning} builds the reasoning path, followed by a detailed introduction of the two verification tasks employed throughout the execution process. As illustrated in Figure~\ref{method}, \textit{Explicit Reasoning Path Planning} organizes the verification process into three stages: Implicit Entity Resolution, Semantic Decomposition, and Integrative Logical Reasoning.

\subsubsection{Implicit Entity Resolution}
The reasoning path begins by identifying implicit entities within the claim. We formalize this process as follows. Let $T = \{t_1, t_2, t_3,\dots, t_m\}$ denote the components of a claim associated with external evidence, where $t_i$ represents a word in the claim and $T$ typically represents a description of an entity. For example, in the claim "The Rookie of The Year in the 1997 CART season drives it in the NASCAR Sprint Cup Series," $T = $ \{"The", "Rookie", "of", "The", "Year", "in", "the", "1997", "CART", "season"\}. Implicit Entity Resolution replaces the entity description $T$ with the concrete entity $T'$ that it denotes.
To implement this stage, we introduce three components $(\mathcal{L}, \mathcal{S}, \mathcal{R})$, corresponding to the Locate, Search, and Replace operations.

\textbf{Locate ($\mathcal{L}$):} This component identifies the span $T$ corresponding to an implicit entity description within a claim $c$. To support the subsequent reasoning steps, the LLM generates a targeted question $Q_l= \mathcal{L}(c)$ about the entity description.
For instance, in Figure \ref{method}, given the description "the band that recorded 'Make Them Gold'", the LLM generates the question: "Which band recorded 'Make Them Gold'?" Specifically, we use the following prompt format: "\texttt{<claim>} Design a question to discover the implicit entity. If no implicit entity is found, print: 'No implicit entity'". 

\textbf{Search ($\mathcal{S}$):} 
Given the generated query $Q_l$, this component accesses external evidence to derive an answer $A_l$ that explicitly contains the identified entity $T'$. This operation is denoted as $A_l = \mathcal{S}(Q_l, \text{Evidence})$.
Specifically, we generate prompts in the following format: "\texttt{<evidence>} Based on the above facts, answer the question. \texttt{<question>}". Subsequently, LLMs will return the answer $A_l$ to the question $Q_l$, and $A_l$ includes the identified implicit entity $T'$. 

\textbf{Replace ($\mathcal{R}$): }This final operation leverages LLMs to replace the original description $T$ in claim $c$ with the entity $T'$ found in $A_l$.
The constructed prompt format is as follows: "\texttt{<Answer>} on the above information, replace the implicit entity and its description in the following expressions with specific names: \texttt{<claim>}". We also adopt dynamic sample selection to provide demonstrations of the replacement process.
Ultimately, we obtain a claim $c' = \mathcal{R}(c, A_l)$ that has undergone the replacement process.

\subsubsection{Semantic Decomposition}

After \textit{Implicit Entity Resolution}, \textit{Semantic Decomposition} verifies the refined claim by examining each of its semantic components individually. This explicit decomposition prevents the LLM from overlooking fine-grained details that become increasingly important in claims requiring more reasoning hops.
Specifically, \textit{Semantic Decomposition} generates sub-questions $Q_s$ for the components of claim $c'$ and obtains the corresponding answers from external knowledge, where $Q_s = \{q_i\}_{i=1}^{n}$ represents a question. For the claim "Greater Swiss Mountain Dog and Harrier are both dog breeds," separate questions can be posed for its complete semantic components, such as "Is the Greater Swiss Mountain Dog a dog breed?" and "Is the Harrier a dog breed?" In Figure \ref{method}, we likewise generate two questions for a claim.
We subsequently verify the answer to each sub-question against the evidence repository.
As in the first stage, we provide the evidence to the LLM as external knowledge.
For each direct question $q_i$, the LLM produces the corresponding answer $a_i, i\in[1,n]$. All the answers are represented as the set $A_s = \{a_1, a_2,...a_n\}$. The specific construction of prompts is as follows: "\texttt{<evidence>} Base on the above facts, \texttt{\texttt{<question>}}".

\subsubsection{Integrative Logical Reasoning} 

Finally, Integrative Logical Reasoning combines the outputs of the preceding stages into a coherent reasoning chain. We manually construct chain-of-thought templates and integrate the outputs from Implicit Entity Resolution and Semantic Decomposition into these templates. The LLM can then review the preceding verification process and derive the final answer solely from the provided content.
Specifically, the chain-of-thought (CoT) template is based on $c'$ and $A_s$ and guides the LLM through the final reasoning process.
The prompt for constructing the CoT is: "Determine whether the following statement is true: <$c'$> True or false? Think step by step: <$A_s$> So, the statement is:". In the end, LLMs provide the answer $y \in\{True, False\}$. At this stage, the LLM focuses on integrating the verified intermediate results without requiring additional external knowledge.

Although explicit reasoning path planning improves the interpretability of the verification process, errors may still arise during individual sub-tasks and propagate to subsequent stages. In particular, the execution pipeline contains two functionally different types of sub-tasks. Evidence-comprehension tasks require the agent to derive an answer from retrieved evidence, whereas instruction-driven reasoning tasks require it to manipulate, transform, or integrate information according to predefined instructions. We therefore introduce two corresponding post-verification mechanisms: \textit{Evidence-Drift Verification} for evidence-comprehension tasks and \textit{Reasoning Reflection Verification} for instruction-driven reasoning tasks.

\subsection{Evidence-Drift Verification}

Evidence-comprehension tasks require the agent to derive an answer directly from the provided evidence. They include, for example, the Search operation in Implicit Entity Resolution and the answer-generation steps in Semantic Decomposition.

To detect evidence drift, we introduce an evidence-free counterpart of the same query. Specifically, given the question $q$, we additionally prompt the LLM to answer $q$ without access to the evidence $e$, relying solely on its parametric memory, denoted as $a^{p} = \mathcal{S}(q, \varnothing)$, where $\varnothing$ indicates that no evidence is supplied. We then compare the evidence-grounded answer $a$ with the parametric answer $a^{p}$. If the two answers converge, i.e., $a \equiv a^{p}$, we cannot rule out the possibility that $a$ was in fact produced from parametric memory rather than genuinely grounded in $e$, in which case we flag the sub-task as a candidate instance of evidence drift.
Once flagged, we explicitly require the agent to re-derive the answer while quoting the specific span of evidence $e^{*} \subseteq e$ that supports its conclusion, formulated as $\hat{a} = \mathcal{S}(q, e, e^{*})$. By forcing the agent to cite $e^{*}$ verbatim, this post-verification step re-anchors the answer to the retrieved evidence rather than the LLMs internal prior, thereby suppressing evidence drift and improving the faithfulness of comprehension.

\subsection{Reasoning Reflection Verification}

Reasoning tasks correspond to the sub-steps in which the agent follows our predefined instructions to resolve the current claim, such as the Locate operation $\mathcal{L}$ in Implicit Entity Resolution and the reasoning step in Integrative Logical Reasoning. Rather than taking the reasoning output at face value, we design a self-reflective post-hoc validation that recasts the produced result as an object to be checked, exploiting the observation that verification capabilities of LLM typically surpass the generative abilities.

Formally, let $x$ denote the input to a reasoning sub-task (e.g., the claim $c'$ or the intermediate answers $A_s$) and $o = \mathcal{F}(x)$ denote the corresponding output produced by the agent, where $\mathcal{F} \in \{\mathcal{L}, \mathcal{R}, \cdot\}$ denotes the reasoning operation being examined. We treat the pair $(x, o)$ as an executable instruction-output pair and construct a verification prompt $\mathcal{V}(x, o)$ that asks the LLM to check whether $o$ correctly and consistently follows from $x$. Crucially, we prepend the verification prompt with an explicit task framing: ``This is part of a fact-checking task, and any error or inconsistency found must be reported.'' This framing shifts the LLM from a generator to a verifier, i.e.,
\begin{equation}
    \hat{o} =
    \begin{cases}
        o, & \text{if } \mathcal{V}(x,o) = \text{consistent}, \\
        \mathcal{F}(x \mid \mathcal{V}(x,o)), & \text{otherwise},
    \end{cases}
\end{equation}
where $\mathcal{V}(x,o) \in \{\text{consistent}, \text{inconsistent}\}$ denotes the verification verdict, and $\mathcal{F}(x \mid \mathcal{V}(x,o))$ regenerates the output given the flagged inconsistency when a flaw is detected. By decoupling verification from generation and explicitly soliciting error reports under the fact-checking framing, this  leverages the superior discriminative ability of LLM to substantially improve the correctness of the agent reasoning.

\begin{table*}[t!]
    \centering
    \caption{Main results (\%). Macro-F1 scores of ReflectFact and baselines on two datasets.
    Results in bold are the best performance.}

    \setlength{\tabcolsep}{3mm}
    \definecolor{groupbg}{RGB}{245,245,245}   
    \definecolor{ReflectFactbg}{RGB}{235,245,255} 
    \renewcommand{\arraystretch}{1.12}

    \begin{tabular}{llccccccc}
    \toprule
        & \multirow{2}{7em}{\textbf{Model}}
        & \multicolumn{4}{c}{\textbf{HOVER}}
        & \multicolumn{3}{c}{\textbf{EX-FEVER}} \\
        \cmidrule(lr){3-6}\cmidrule(lr){7-9}
        & & \textbf{2-hop} & \textbf{3-hop} & \textbf{4-hop} & \textbf{Total}
          & \textbf{2-hop} & \textbf{3-hop} & \textbf{Total} \\
    \midrule

    \rowcolor{groupbg}
    \multicolumn{9}{l}{\textbf{\uppercase\expandafter{\romannumeral1}. Vanilla LLM}} \\
        & FLAN-T5 \cite{chung2022scaling}
        & 64.23 & 56.39 & 54.19 & 56.41 & 72.42 & 61.72 & 67.64\\
        & Qwen3 \cite{yang2025qwen3}
        & 76.09 & 74.48 & 59.74 & 71.48 & 79.61 & 73.04 & 73.93\\
        & GPT-4o-mini \cite{openai2024gpt4ocard}
        & 76.72 & 73.10 & 68.89 & 74.11 & 80.10 & 72.11 & 74.45\\
    \midrule

    \rowcolor{groupbg}
    \multicolumn{9}{l}{\textbf{\uppercase\expandafter{\romannumeral2}. Inference augmented model}} \\
        & Scandi-NLI \cite{ScandiNLI}
        & 72.98 & 64.85 & 57.66 & 65.35 & 65.66 & 59.48 & 62.75\\
        & DeBERTaV3-NLI \cite{laurer2024less}
        & 73.93 & 63.88 & 51.69 & 64.20 & 69.25 & 61.83 & 66.10\\
        & ProgramFC \cite{pan2023fact}
        & 76.32 & 68.85 & 64.96 & 69.31 & 83.65 & 78.24 & 80.98\\

    \midrule

    \rowcolor{groupbg}
    \multicolumn{9}{l}{\textbf{\uppercase\expandafter{\romannumeral3}. Agent-based Fact Verification}} \\
        & HiSS \cite{zhang2023towards}
        & 75.45 & 65.10 & 66.02 & 68.21 & 82.28 & 74.91 & 78.61\\
        & Factcheck-GPT \cite{wang-etal-2024-factcheck}
        & 75.33 & 67.84 & 67.31 & 71.34 & 82.81 & 75.10 & 78.96\\
        & StepByStepFV \cite{vladika2025step}
        & 77.86 & 73.42 & 69.87 & 74.53 & 82.34 & 77.11 & 79.68\\
        & BiDeV \cite{10.1609/aaai.v39i1.32034}
        & 80.08 & 74.57 & 69.95 & 75.19 & 83.48 & 76.38 & 79.66\\

        \rowcolor{ReflectFactbg}
        & \textbf{ReflectFact}
        & \textbf{83.33} & \textbf{76.91} & \textbf{73.74} & \textbf{78.51}
        & \textbf{86.68} & \textbf{80.55} & \textbf{83.76}\\

    \bottomrule
    \end{tabular}
    \label{chart:ans}
    \vspace{-0.5em}
\end{table*}

\section{Experiments}
\textbf{Datasets.} To better align with the requirements of real-world fact verification, our experiments are conducted on two multi-hop fact verification datasets, HOVER \cite{jiang2020hover} and EX-FEVER \cite{ma2023ex}. HOVER is a multi-hop dataset derived from English Wikipedia articles. We utilize the validation set of HOVER for evaluation, which consists of 4,000 claims. The claims in HOVER require evidence from up to four Wikipedia articles to determine whether the claims are true. EX-FEVER involves 2-hop and 3-hop reasoning, claims are created by summarizing and modifying information from hyperlinked Wikipedia documents. We evaluate the model using the test set of EX-FEVER. To maintain consistency with HOVER, we remove data with the NEI label. After that, the number of claims is 4,071.

\textbf{Baselines.} We use ten baselines for comparison with ReflectFact, which can be categorized into three categories. (\uppercase\expandafter{\romannumeral1}). 
\textbf{Vanilla LLM}, we employ both open-source and closed-source LLMs as verifiers. For the open-source model, we use Flan-T5 \cite{chung2022scaling}, a unified Text-to-Text Transformer. We create a prompt by combining the claim and evidence and obtain the classification result through the Text-to-Text approach. For the closed-source and open-source models with stronger backbones, we use GPT-4o-mini \cite{openai2024gpt4ocard} and Qwen3 \cite{yang2025qwen3}, which are queried with the same prompt to directly produce a verdict for the claim given its evidence. (\uppercase\expandafter{\romannumeral2}). 
\textbf{Inference augmented model}, ScandiNLI \cite{ScandiNLI} fine-tunes nb-bert-large for Natural Language Inference (NLI). DeBERTaV3-NLI \cite{laurer2024less} fine-tunes the DeBERTaV3 on annotations from FEVER and four NLI datasets. ProgramFC \cite{pan2023fact} uses a shared library of specialized functions to reason with the help of codex and Flan T5. (\uppercase\expandafter{\romannumeral3}). 
\textbf{Agent-based Fact Verification}, based on the GPT-4o-mini engine. Rather than producing a verdict in a single pass, these agents autonomously plan a sequence of processing steps. Compared with directly prompting an LLM for the final label, this agentic paradigm offers a more transparent and flexible reasoning process for handling complex multi-hop claims.
The HiSS \cite{zhang2023towards} uses LLMs to partition the claim and determine the truth of each subpart. 
Factcheck-GPT \cite{wang-etal-2024-factcheck} is a fine-grained system for fact-checking. 
Besides, we use BiDeV \cite{10.1609/aaai.v39i1.32034}, which integrates multiple role-playing LLM agents that mutually defuse and cross-validate each other to fact-check claims. For fairness, we replace the backbone of every agent-based baseline with GPT-4o-mini and impose a constraint that the model cannot access the web.

\textbf{Metrics.}
Following prior multi-hop fact-verification studies~\cite{wang-etal-2024-factcheck,10.1609/aaai.v39i1.32034}, we adopt Macro-F1 as the primary evaluation metric to jointly assess performance on supported and refuted claims. Specifically, we report hop-wise Macro-F1 scores to examine how verification accuracy varies with increasing reasoning depth, together with the overall Macro-F1 on each dataset.

\textbf{Implementation Details.} For the fine-tune models, we use cross-entropy loss and the AdamW optimizer with a learning rate of 1e-5. We employ gpt-4o-mini as the backbone for the agent-based fact verification. To stabilize the output of model, we use a greedy decoding strategy, setting the temperature to 0. We also use the FLAN-T5-XL 3B as the backbone of the T5 module.

\subsection{Comparison with State-of-the-art Methods}
As shown in Table \ref{chart:ans}, we demonstrate the performance of ReflectFact and baselines. Furthermore, we can observe the following findings:

\textbf{Stronger large language models and better reasoning paradigms help improve fact-checking.}
Although Scandi-NLI and DeBERTaV3-NLI are specifically fine-tuned on NLI annotations, they still lag behind Vanilla LLMs such as GPT-4o-mini and Qwen3 by a large margin, e.g., DeBERTaV3-NLI trails GPT-4o-mini by 9.91\% and 8.35\% on HOVER and EX-FEVER, respectively. We attribute this to the strong reasoning capability of fine-tuned NLI models, despite the fact that their underlying base models are not sufficiently powerful. Notably, ProgramFC also benefits from this property, as it relies on an LLM to generate reasoning programs. Nevertheless, even the strongest Vanilla LLM still falls short of ReflectFact by 4.40\% and 9.31\% on both datasets, indicating that LLMs still require explicit guidance to reason reliably over multiple hops.

\textbf{Compared with directly using LLMs, agent-based fact verification methods exhibit a higher performance ceiling when handling complex claims.}
On simpler 2-hop claims, the strongest Vanilla LLM, GPT-4o-mini, already performs competitively, but this advantage fades as the number of hops grows. On the most challenging 4-hop claims, the best agent-based baselines, BiDeV and StepByStepFV, reach 69.95\% and 69.87\%, surpassing GPT-4o-mini (68.89\%), while HiSS and Factcheck-GPT, which merely partition the claim without an explicit verification workflow, still fall behind. This indicates that decomposing a claim and progressively resolving it through an agent workflow is more effective than a single LLM call for long-chain reasoning. Furthermore, ReflectFact demonstrates even more substantial improvements, outperforming the strongest agent-based baseline and Vanilla LLM by 3.79\% and 4.85\% on 4-hop claims, respectively.

\textbf{The self-reflective post-verification unlocks the full potential of the agent, enabling ReflectFact to achieve the best overall results.} HiSS \cite{zhang2023towards} and Factcheck-GPT \cite{wang-etal-2024-factcheck} only focus on partitioning the claim without further examining whether the resulting sub-answers and reasoning steps are actually reliable, while BiDeV \cite{10.1609/aaai.v39i1.32034} relies solely on multi-agent cross-validation without revisiting its own evidence usage. Consequently, even the strongest agent baseline, BiDeV, still trails ReflectFact by 3.32\% and 4.10\% in overall performance on HOVER and EX-FEVER, respectively. In contrast, by explicitly re-examining evidence grounding and reasoning validity after each execution step, ReflectFact corrects flawed intermediate outputs before they propagate to the final verdict, allowing it to consistently surpass all agent-based baselines and achieve the best overall results on both datasets.

\subsection{Ablation Study}

In this section, we conduct an ablation analysis of the ReflectFact, investigating the impact of two verification tasks:  Evidence-Drift Verification, Reasoning Reflection Verification. The results are shown in Table \ref{chart:Ablation}.

\begin{table}[t]
\small
    \centering
    \caption{Ablation results (\%). Macro-F1 scores for ReflectFact and the ablation setting.}
    \setlength{\tabcolsep}{2.5mm}{
    \begin{tabular}{lccccc}
        \toprule
        \multirow{2}{4em}{} & \multicolumn{3}{c}{\textbf{HOVER}} & \multicolumn{2}{c}{\textbf{EX-FEVER}} \\ 
        \cmidrule(lr){2-4}\cmidrule(lr){5-6}
         & \textbf{2-hop} & \textbf{3-hop} & \textbf{4-hop} & \textbf{2-hop} & \textbf{3-hop} \\
        \midrule 
        w$\backslash$o RRV & 80.68 & 73.77 & 69.98 & 84.04 & 76.68 \\ 
        w$\backslash$o EDV & 82.69 & 76.02 & 72.57 & 84.21 & 77.84 \\ 
        ReflectFact & 83.33 & 76.91 & 73.74 & 86.68 & 80.55 \\ 
        \bottomrule
    \end{tabular}}
    \label{chart:Ablation}
    \vspace{-0.5em}
\end{table}



\textbf{Effect of Evidence-Drift Verification (EDV).}
Removing EDV causes a moderate but consistent performance drop that widens as the number of hops increases. Since EDV detects and corrects evidence drift, i.e., cases where the agent answer coincides with its evidence-free prediction, its removal leaves such drifted answers uncorrected, and these early comprehension errors are more likely to propagate into later reasoning steps on longer claims.

\textbf{Effect of Reasoning Reflection Verification (RRV).}
Removing RRV leads to a substantially larger degradation than removing EDV. Without RRV, the outputs of reasoning sub-steps, such as locating implicit entities or constructing the final logical chain, are accepted at face value rather than being re-examined from a perspective of verifier. This confirms that recasting the agent own reasoning output as an object to be verified is the primary source of ReflectFact robustness, particularly on claims that demand longer chains of reasoning.

\begin{figure*}[t!]
	\centering
	\subfigure[ReflectFact and baselines based on  GPT-4o-mini] {\includegraphics[width=.48\textwidth]{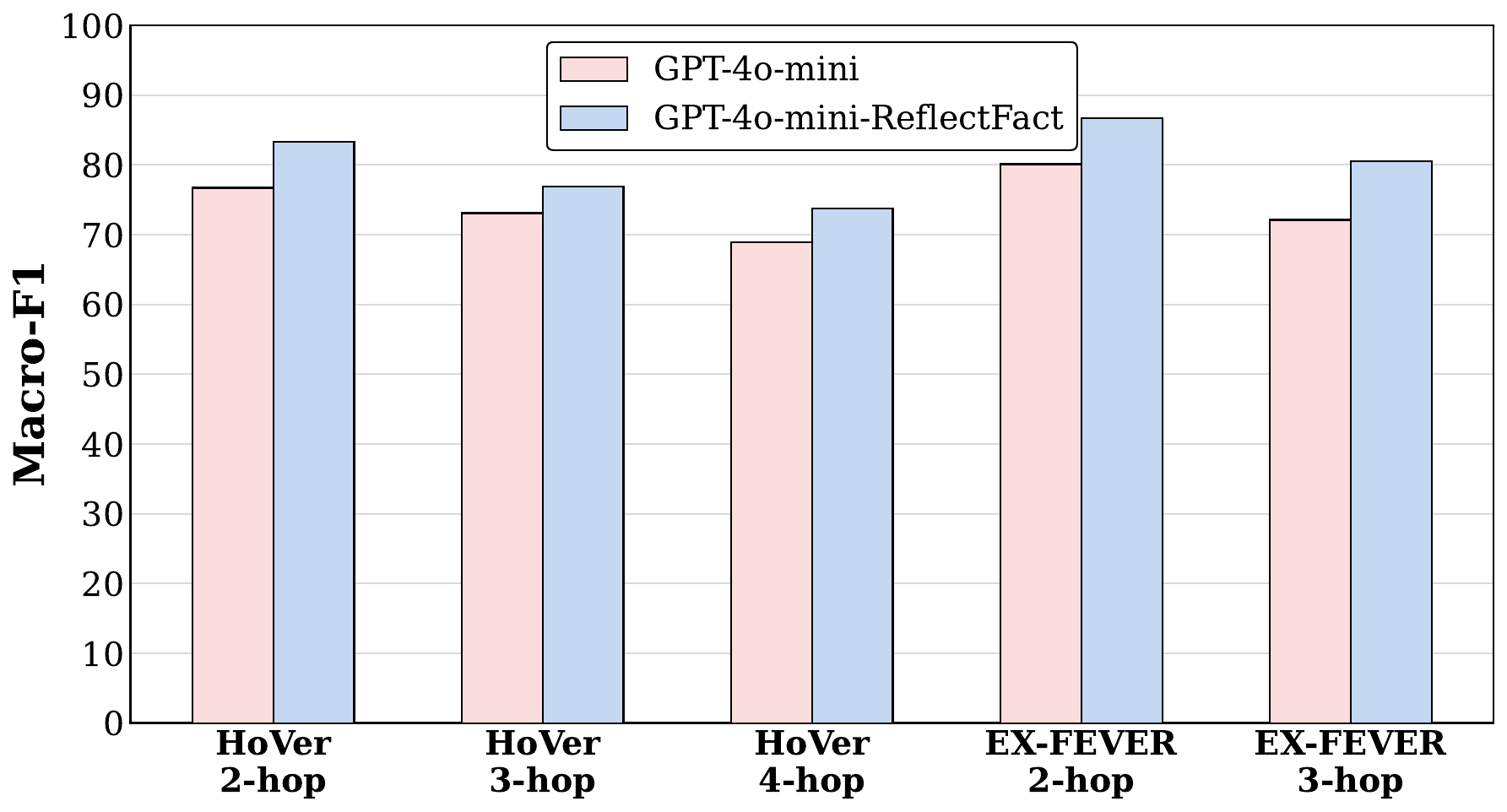}}
	\subfigure[ReflectFact and baselines based on  Qwen3-8B] {\includegraphics[width=.48\textwidth]{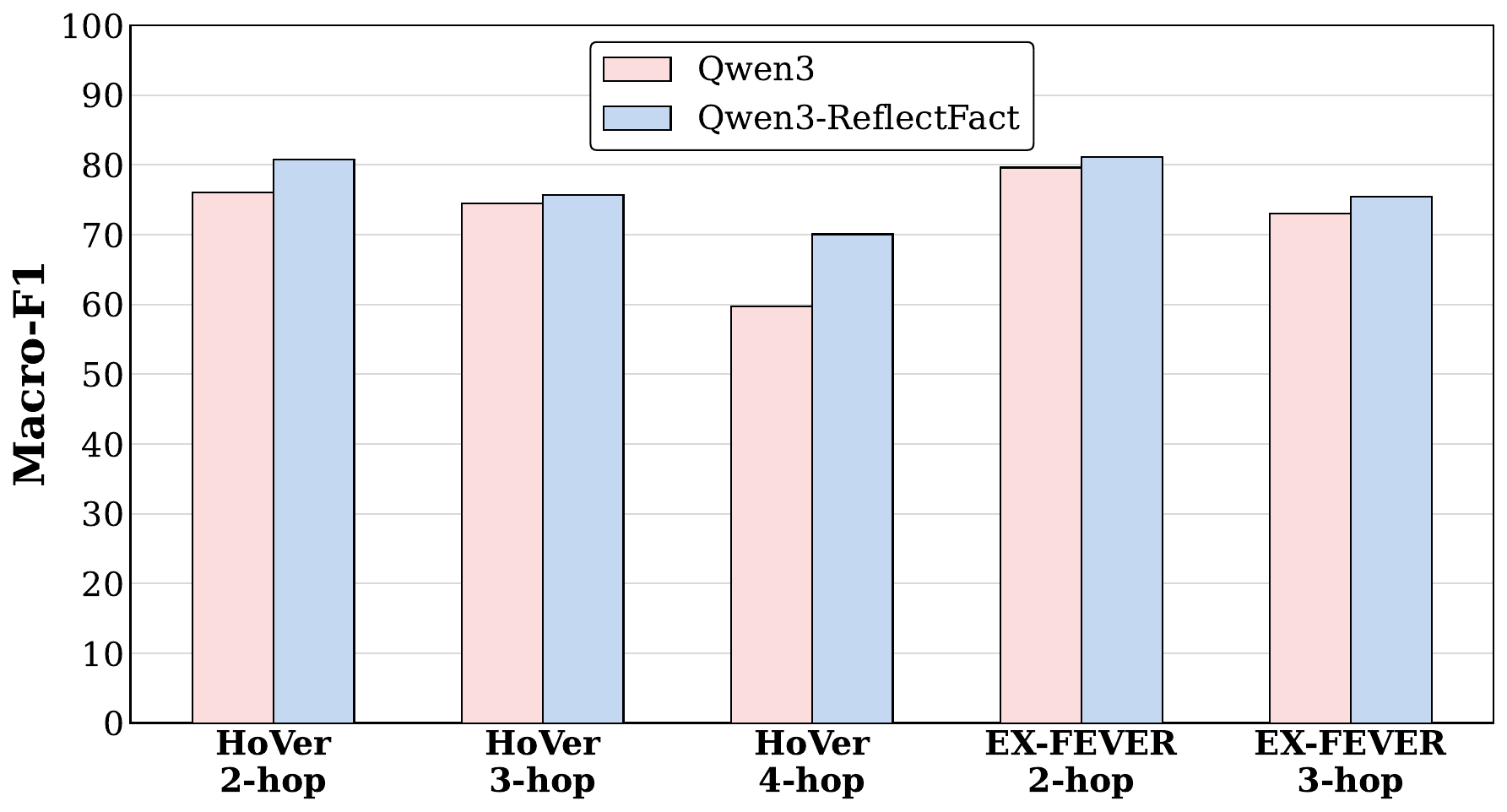}}
	\caption{Macro-F1 scores of different LLMs based ReflectFact and baselines on two datasets.}
	\label{dLLMs}
    \vspace{0.5em}
\end{figure*}

\subsection{Model Generalizability}
\label{exp:sca}
ReflectFact is designed as a general, model-agnostic framework for enhancing the
multi-hop fact-verification ability of different LLM backbones. To assess its
generalizability, we instantiate it on two representative backbones of different
scales and access levels: the proprietary GPT-4o-mini \cite{openai2024gpt4ocard}
and the open-source Qwen3-8B \cite{yang2025qwen3}. For each backbone, we compare ReflectFact against its standard prompting baseline under an identical evaluation protocol.
As shown in Figure~\ref{dLLMs}(a) and Figure~\ref{dLLMs}(b), ReflectFact delivers
consistent Macro-F1 improvements across both backbones and all settings on the
HOVER and EX-FEVER benchmarks. The gains hold for both the proprietary and the
open-source model, indicating that the benefits of ReflectFact are not tied to a
particular model scale or family. More notably, its advantage becomes more
pronounced as the reasoning depth increases: while the baselines degrade sharply
on instances requiring more hops, ReflectFact substantially alleviates this
decline and attains its largest improvements precisely in the most challenging
multi-hop settings. These results confirm that ReflectFact robustly and generally
strengthens the multi-hop fact-verification reasoning of diverse LLMs.

\subsection{Interpretability Analysis}
\label{exp:exp}

EX-FEVER dataset provides golden explanations for the label, which is a textual explanation that describes the minimally sufficient information in each hop to verify a claim. To verify the interpretability of the ReflectFact, we conducted experiments using the golden explanations data from the EX-FEVER dataset. We consider that the constructed CoT, as an intermediate result of ReflectFact, is accessible and can provide explanations for the judgments. We use the ROUGE metrics to measure the matching degree between the constructed CoT and the golden explanation. Ultimately, we found that Constructed CoT has an explanatory role. The experimental results are shown in Table \ref{chart:Exp}. We compared ReflectFact with three models: MDR \cite{xiong2021answering}, BERT \cite{devlin2018bert}, and GPT \cite{brown2020language}, as used in the study by \cite{ma2023ex}. Although our model is not explicitly designed for generating explanations, it outperforms three models on average by 5.36\% and 7.96\% in Rouge-1 and Rouge-L metrics. 


\begin{table}[t!]
    \centering
    \caption{Results of different models generating explanations (\%). Results in bold are the best performance.}
    \resizebox{\columnwidth}{!}{ 
    \setlength{\tabcolsep}{3mm}{
    \begin{tabular}{ccccc}
        \toprule
        & MDR & BERT-based & GPT & ReflectFact\\
        \midrule
        Rouge-1& 54.88 & 46.88 & 52.28 & \textbf{56.71}\\
        Rouge-2& \textbf{41.34} & 32.80 & 33.74 & 39.87\\
        Rouge-L& 49.42 & 35.52 & 48.13 & \textbf{52.32}\\
        \bottomrule
    \end{tabular}}}
    \label{chart:Exp}
\end{table}

\subsection{Error Type Analysis}

\begin{figure}[htbp]
    \centering
    \includegraphics[width=0.49\textwidth]{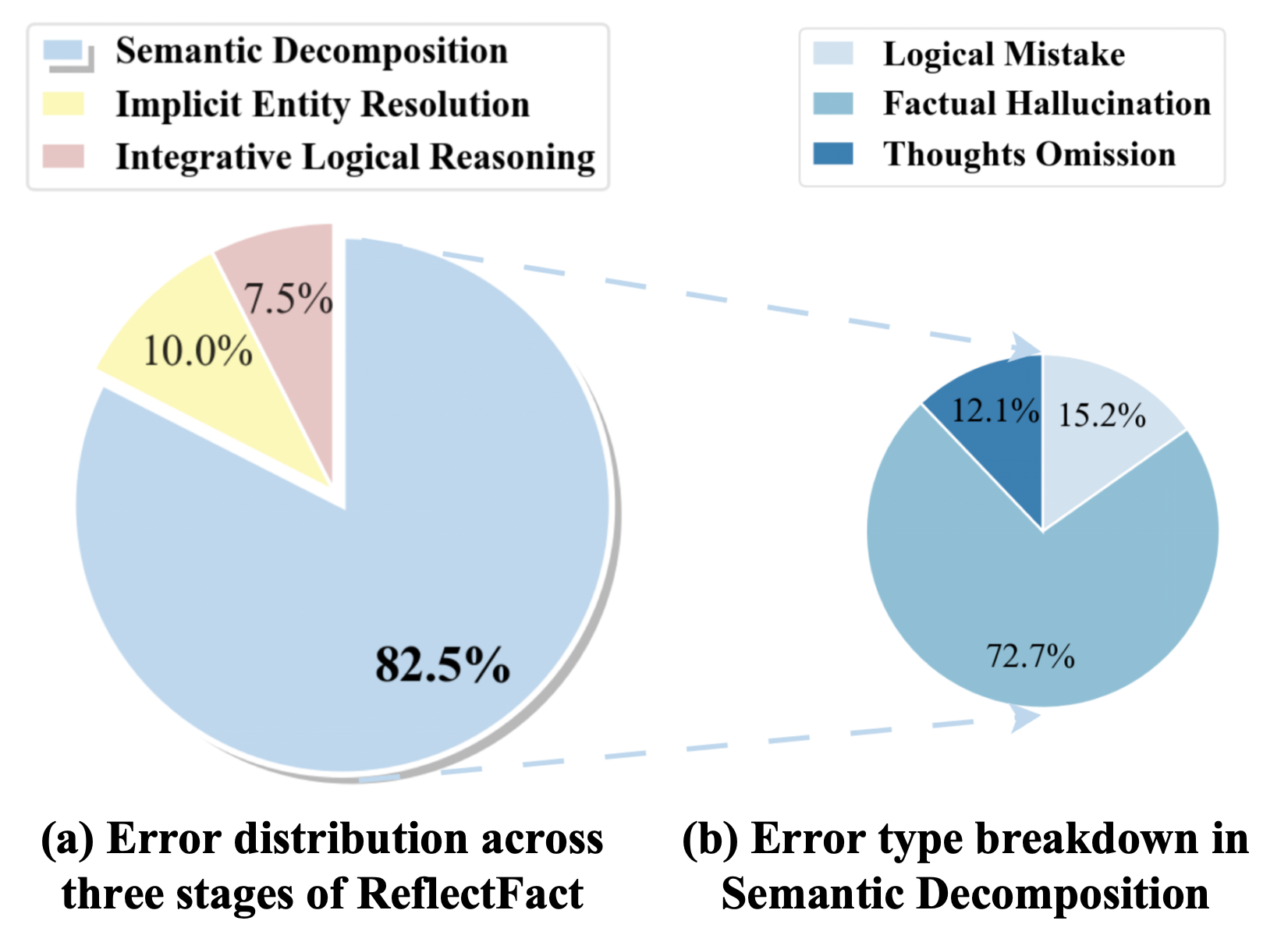}
    \caption{Error analysis of ReflectFact on 40 samples. }
    \label{fig:error}
\end{figure}

To better understand the behavior of our ReflectFact and facilitate future research, we randomly sampled 40 examples that are generated by ReflectFact. As shown in Figure \ref{fig:error}(a), we categorize the causes of errors into three types: 1) \textit{logical mistake}, 2) \textit{factual hallucination}, and 3) \textit{thoughts omission}. 
We find that the errors corresponding to the three stages are distributed as 10\%, 82.5\%, and 7.5\%, respectively. 
Consequently, we further analyzed the types of errors occurring in the Semantic Decomposition.
As shown in Figure \ref{fig:error}(b), 72.7\% of the errors are concentrated in hallucinations generated by LLM itself. Meanwhile, although decomposing sub-information enhances the logical reasoning of LLMs, it also ignores longer dependencies in the text, resulting in some degree of performance loss. 

\section{Conclusion}
In this paper, we identify two critical limitations of existing agent-based fact-checking methods, namely their susceptibility to objective conflicts and knowledge conflicts.
To address these issues, we propose ReflectFact, a novel self-reflective agent framework for multi-hop fact verification. ReflectFact employs Evidence-Drift Verification and Reasoning Reflection Verification to post-verify comprehension and reasoning sub-tasks, respectively, thereby correcting flawed intermediate outputs before they propagate to the final verdict.
Extensive experiments on two datasets show that ReflectFact consistently achieves state-of-the-art performance, and further analyses confirm that it generalizes well across different backbones while offering strong interpretability.
\bibliography{aaai2027}


\end{document}